\documentclass[conference,10pt,a4paper]{IEEEtran}
\usepackage{graphicx}
\usepackage{epstopdf}
\usepackage{subfigure}
\usepackage{multirow}
\usepackage[english]{babel}
\usepackage[utf8]{inputenc}
\usepackage{algorithm}
\usepackage[noend]{algpseudocode}
\usepackage{color}
\usepackage{subfigure} % Written by Steven Douglas Cochran
\usepackage{url}       % Written by Donald Arseneau
\usepackage{amsmath}   % From the American Mathematical Society
\usepackage{array}

\begin{document}

\title{Bundle Adjustment Revisited}

% avoiding spaces at the end of the author lines is not a problem with
% conference papers because we don't use \thanks or \IEEEmembership

\author{Yu Chen{1}, Yisong Chen{1}, Guoping Wang{1} \\
{1} Peking University, Department of Computer Science and Technology, \\ Graphics and Interactive Lab, Beijing, China
%\authorblockA{\authorrefmark{2}Name of Institution/Department, City, Country}
}

%\specialpapernotice{(Invited Paper)}

\maketitle

\begin{abstract}
3D reconstruction has been developing all these two decades, from moderate to medium size and to large scale. It's well known that bundle adjustment plays an important role in 3D reconstruction, mainly in Structure from Motion(SfM) and Simultaneously Localization and Mapping(SLAM). While bundle adjustment optimizes camera parameters and 3D points as a non-negligible final step, it suffers from memory and efficiency requirements in very large scale reconstruction. In this paper, we study the development of bundle adjustment elaborately in both conventional and distributed approaches. The detailed derivation and pseudo codes are also given in this paper. 

\end{abstract}

% no key words

\section{Introduction}
% no \PARstart
Bundle adjustment plays an important role in geodesy and 3D reconstruction(both SLAM and SfM), but it was deemed as solved until recently. Bundle adjustment constitutes a core component in most state-of-the-art multi-view geometry systems and is typically invoked as a final refinement stage to approximate initial scene estimates as well as a means for removing drift in incremental reconstructions. The Levenberg-Marquardt algorithm has proven to be the most successful method for solving this formulation, as it is simple to implement, robust to initialization, and its framework makes it very amenable to taking advantage of the forms of sparsity that typically arise in multi-view geometry problem. Each step of this algorithm produces an estimate of the parameters that improves upon the previous and the resulting series of iterates can be shown to converge to a local minimum of the objective function at hand. 

As a result, a number of approaches to bundle adjustment have been proposed in the last decade. These approaches divide into two groups: the first branch focuses on making the bundle adjustment algorithm as efficient as possible, while the second focuses on reducing the size or frequency of invocation of individual bundle adjustment.

Bundle adjustment is the key to refining a visual reconstruction to produce jointly optimal 3D structure and viewing parameter estimates. Optimal means that the parameter estimates are found by minimizing some cost function that quantifies the model fitting error, and jointly that the solution is simultaneously optimal with respect to both structure and camera variations. The name refers to the 'bundle' of light rays leaving each 3D feature and converging on each camera center, which are 'adjusted' optimally with respect to both feature and camera positions\cite{DBLP:conf/iccvw/TriggsMHF99}.

Since the basic mathematics of the bundle adjustment problem are well understood\cite{DBLP:conf/iccvw/TriggsMHF99} and several implementation-aimed works are given\cite{DBLP:journals/toms/LourakisA09, DBLP:conf/cvpr/WuACS11}, we can use some open-source software packages easily\cite{DBLP:journals/toms/LourakisA09, DBLP:conf/eccv/Zach14, DBLP:conf/eccv/AgarwalSSS10, DBLP:conf/cvpr/WuACS11, ceres-solver}. However, bundle adjustment is still a bottleneck in large-scale Structure from Motion because of matrix storage and frequent matrix manipulation\cite{DBLP:conf/eccv/FrahmGGJRWJDCL10, DBLP:conf/3dim/Wu13, DBLP:conf/accv/MoulonMM12, DBLP:conf/cvpr/SchonbergerF16, DBLP:conf/mm/SweeneyHT15, DBLP:conf/eccv/WilsonS14, DBLP:journals/pami/CrandallOSH13}. The naive LM algorithm requires $O((m+n)^3$ operations for each iteration, and memory on the order of $O(mn(m+n))$. However, exploiting matrix structure and using the Schur complement approach, the number of arithmetic operations can reduced to $O(m^3+mn)$, and memory use to $O(mn)$. Further reduction can be achieved by exploiting secondary sparse structure\cite{DBLP:conf/bmvc/Konolige10}. The conjugate gradient approaches in \cite{DBLP:conf/eccv/AgarwalSSS10, DBLP:conf/eccv/ByrodA10} can reduce the time complexity to $O(m)$ per iteration, making it essentially linear in the number of cameras. \cite{DBLP:journals/toms/LourakisA09} uses dense Cholesky decomposition to solve the normal equation directly, and it needs much memory to store Jacobi and the reduced camera system, thus it's not suitable for large scale reconstruction work. \cite{DBLP:conf/eccv/AgarwalSSS10} uses preconditioning on either hessian or schur complement and avoid the explicit storage of Jacobi, and the preconditioned conjugate gradient approaches\cite{DBLP:conf/eccv/ByrodA10, DBLP:conf/bmvc/ByrodA09, DBLP:conf/iccv/JianBD11} makes the procedure of solving the reduced camera system more stable because it has lower condition number than solving the Gauss-Newton\cite{DBLP:books/sp/NocedalW99} or Levenberg-Marquardt\cite{levenberg-marquardt} problem directly. 

Robust approaches\cite{DBLP:conf/icip/AravkinSMNB12, DBLP:journals/mta/CaoLJLL17} are typically used to protect world point and camera parameter estimates from effect of outliers, which for BA are incorrect point correspondences that have gone undetected.

The development of hardware, especially GPU(Graphics Processing Unit), makes some works concentrate on the implementation of parallel bundle adjustment\cite{DBLP:conf/cvpr/WuACS11} and to accelerate the reconstruction. 

However, when the data scale gets larger, e.g, city-scale reconstruction\cite{DBLP:conf/iccv/ZhangZFQ17, Zhu2017Parallel, DBLP:conf/cvpr/ZhuZZSFTQ18}, the approaches discussed above cannot meet the memory and efficiency requirements. Thus, some works\cite{DBLP:conf/iccv/ZhangZFQ17, DBLP:conf/iccvw/RamamurthyLAPV17, DBLP:conf/cvpr/ErikssonBCI16} concentrate on performing bundle adjustment in distributed manner. The limitation of memory can be avoided once we have enough computers. Besides, a large improvement in both accuracy and efficiency would happen. While\cite{DBLP:conf/cvpr/ErikssonBCI16} uses Douglas-Rachford splitting methods to split the conventional bundle adjustment problem into distributed manner, \cite{DBLP:conf/iccv/ZhangZFQ17, DBLP:conf/iccvw/RamamurthyLAPV17} implements \textit{ADMM}(Alternating Direction Method of Multipliers) to transform the original problem into a distributed one.

The main purposes of this paper are two folds:
\begin{itemize}
    \item To give a detailed investigation and derivation of bundle adjustment problem, in both theoretical and practical levels.
    \item To show the development of bundle adjustment in parallel mode, and to give a future direction of bundle adjustment.
\end{itemize}

Our paper organizes as follows: we first introduce the projection camera model and camera distortion in Section II, then we show how to perform bundle adjustment with the introduced camera model. In Section III, we give the derivation of conventional bundle adjustment in detail, and the distributed bundle adjustment algorithm is given in Section IV. Finally, we make a conclusion of our work.

\section{Projection Camera Model}

Bundle adjustment describes the sum of errors between the measured pixel coordinates $u_{ij}$ and the re-projected pixel coordinates. The re-projected pixel coordinates are computed by structure(3D points coordinates in world frame) and camera parameters. Thus it is essential to figure out the re-projection process before we get deep into bundle adjustment. We will introduce the pinhole camera model in this section first.

\subsection{Pinhole Camera}
\begin{figure}[ht]
    \centering
    \includegraphics[scale=0.36]{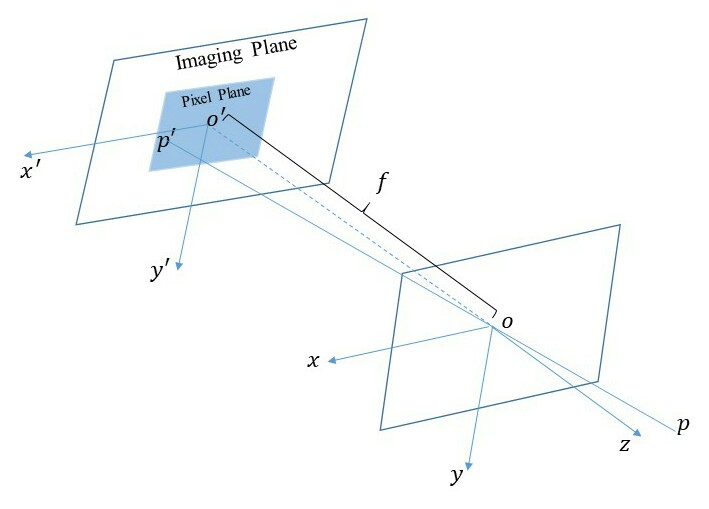}
    \caption{Pinhole camera model: $p$ is a 3D point located in camera frame, $o$ is the camera center, $f$ is the focal length, $p$ is projected into the imaging plane on a pinhole camera model.}
    \label{fig:pinhole_camera_model}
\end{figure}

The pinhole camera model is depicted in Figure.\ref{fig:pinhole_camera_model}, the coordinates of $p$ is $(X,Y,Z)^T$ in camera frame $o-xyz$, the focal length of the camera is $f$, the camera center $o$ is deemed as a pinhole, and $p$ is projected into the imaging plane $o^{'}-x^{'}y^{'}$, and represented by 2D point $p^{'}$.

\begin{figure}
    \centering
    \subfigure[similar triangle in $x-z$ axis] {
        \begin{minipage}{0.8\linewidth}
        \includegraphics[scale=0.5]{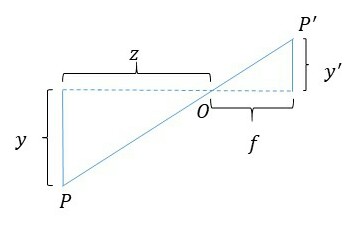}\vspace{5pt}
        \end{minipage}
    }
    \subfigure[similar triangle in $y-z$ axis] {
        \begin{minipage}{0.6\linewidth}
        \includegraphics[scale=0.5]{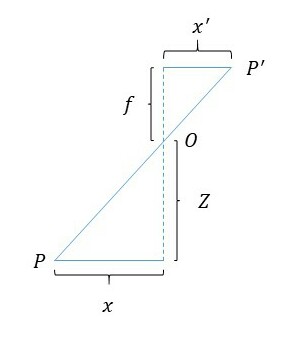}
        \end{minipage}
    }
    \caption{Similar Triangle}
    \label{fig:similar_triangle}
\end{figure}

Based on similar triangle theory(as shown in Figure.\ref{fig:similar_triangle}), we can obtain:
\begin{equation}
\label{similar triangle}
    \frac{f}{Z} = -\frac{X^{'}}{X} = -\frac{Y^{'}}{Y}
\end{equation}

To simplify the model, we can put the imaging plane in front of camera, along with the 3D point, just like Figure.\ref{fig:imaging_plane}:

\begin{figure}
    \centering
    \subfigure[The true imaging plane] {
        \begin{minipage}{0.5\linewidth}
        \includegraphics[scale=0.4]{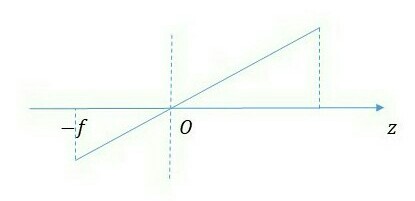}
        \end{minipage}
    }
    \subfigure[The symmetry imaging plane] {
        \begin{minipage}{0.5\linewidth}
        \includegraphics[scale=0.4]{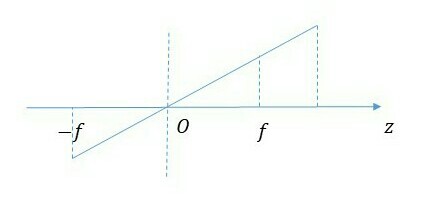}
        \end{minipage}
    }
    \caption{Imaging Plane: (a) is the real imaging plane, (b) is the symmetry imaging plane that we put the imaging plane in (a) in front of camera.}
    \label{fig:imaging_plane}
\end{figure}

Then (\ref{similar triangle}) simplified into $\frac{f}{Z} = \frac{X^{'}}{X} = \frac{Y^{'}}{Y}$, that is
\begin{equation}
\label{inhomo_similar_triangle}
\left\{
    \begin{aligned}
        X^{'} = f \cdot \frac{X}{Z}\\
        Y^{'} = f \cdot \frac{Y}{Z}
    \end{aligned}
\right.
\end{equation}
However, we can only get the coordinate in imaging plane, and, we obtain the pixel coordinate in pixel plane actually. Suppose that there is a pixel plane fixed in the imaging plane $ouv$, the coordinate of point $p^{'}$ in pixel plane is $(\mu, \nu)^T$. Pixel frame and imaging plane related with a scale and translation. Assume pixel coordinate scaled by $\alpha$ in axis $u$, and scaled by $\beta$ in axis $v$, the translation to origin is $(c_x, c_y)^T$. Then the relation between coordinates in imaging plane and pixel plane is:
\begin{equation}
\label{relation_pixel_imaging}
\left\{
\begin{aligned}
    \mu = \alpha X^{'} + c_x\\
    \nu = \beta Y^{'} + c_y
\end{aligned}
\right.
\end{equation}
By inserting (\ref{inhomo_similar_triangle}) into (\ref{relation_pixel_imaging}) and setting $f_x = \alpha \cdot f$, $f_y = \beta \cdot f$, we can obtain:
\begin{equation}
\label{relation_pixel_imaging2}
\left\{
\begin{aligned}
    \mu = f_x \cdot \frac{X}{Z} + c_x\\
    \nu = f_y \cdot \frac{Y}{Z} + c_y
\end{aligned}
\right.
\end{equation}
In usual, (\ref{relation_pixel_imaging2}) is written in matrix format:
\begin{equation}
\label{projection_model}
    \left[
    \begin{array}{c}
         \mu \\
         \nu \\
         1
    \end{array}
    \right] = 
    \left[
    \begin{array}{ccc}
         f_x & 0 & c_x  \\
         0 & f_y & c_y  \\
         0 & 0 & 1
    \end{array}
    \right]
    \left[
    \begin{array}{c}
         \frac{X}{Z} \\
         \frac{Y}{Z} \\
         1
    \end{array}
    \right] = 
    \frac{1}{Z}KP
\end{equation}
where $K$ is called $calibration matrix$ or $internal matrix$.

\begin{figure}
    \centering
    \includegraphics[scale=0.5]{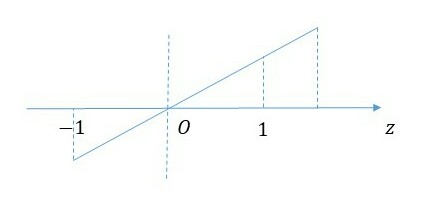}
    \caption{Normalized Imaging Plane}
    \label{fig:normalized_imaging_plane}
\end{figure}

In the equation above, the 3D coordinate of $P$ is located in camera frame because of camera motion, and, we could only obtain its coordinate in world frame $P_w$ in practice. So we have to transform $P_w$ into camera frame before we make use of it:
\begin{equation}
    p_{\mu \nu} = 
    \left[
    \begin{array}{c}
         \mu  \\
         \nu  \\
         1
    \end{array}
    \right] = 
    \frac{1}{Z} \cdot
    K(RP_w + t)
\end{equation} 
where $P_c = \frac{1}{Z}(RP_w + t)$ is called the \textit{normalized coordinate}. It located in front of the plane of camera where $z = 1$, and the plane is called \textit{normalized plane} in Figure.\ref{fig:normalized_imaging_plane}.

\subsection{Camera Distortion}
To achieve a better image formation effect, the optical lens is usually added in front of the camera, thus the light forecasting may be affected during image formation. There are mainly two causes:
\begin{itemize}
    \item the path of ray passing affected by optical lens
    \item the optical lens are not parallel strictly with image formation plane during camera assembling
\end{itemize}

\subsubsection{Radial Distortion}
The distortion caused by the optical lens itself is called \textit{radial distortion}. The camera lens makes a straight line become a curve in an image, and it becomes more clear when getting closer to the border of the image. Radial distortion is usually divided into \textit{barrel distortion} and \textit{pillow distortion}. For radial distortion, we can correct it by using a polynomial function that relates with the distance to the image center. The equation of radial distortion correction is given by: (\ref{radial_distort_eqa})
\begin{equation}
\label{radial_distort_eqa}
\left\{
\begin{aligned}
    x_c = x(1 + k_1r^2 + k_2r^4 + k_3r^6)\\
    y_c = y(1 + k_1r^2 + k_2r^4 + k_3r^6)
\end{aligned}
\right.
\end{equation}
where $(x, y)^T$ is the coordinate before correction, and $(x_c, y_c)^T$ is the coordinate after correction, and $r = x^2 + y^2$. Note that, both $(x, y)^T$ and $(x_c, y_c)^T$ are located in the normalized image plane.

\subsubsection{Tangential Distortion}
As mentioned at the beginning of this section, the tangential distortion is caused by the not strictly parallel between optical lens and image plane. In tangential distortion, we can use two more parameters $p_1$, $p_2$ for correction. The equation of tangential distortion correction is given by (\ref{tangential_distort_eqa})
\begin{equation}
\label{tangential_distort_eqa}
\left\{
    \begin{aligned}
        x_c = x + 2p_1xy + p_2(r^2 + 2x^2)\\
        y_c = y + p_1(r^2 + 2y^2) + 2p_2xy
    \end{aligned}
\right.
\end{equation}

\subsubsection{Performing Bundle Adjustment with Camera Model}
With the discussion above, we can make a summarize of the re-projection process, and it is given by algorithm\ref{reprojection_alg}.  

\begin{algorithm}
\caption{Structure Re-projection Algorithm}
\label{reprojection_alg}
\begin{algorithmic}[1]
    \Require 3D points set(structure) $\mathcal{X} = \{X_i^{world}\}, i \in (1, n)$ in world frame,
    camera parameters $\mathcal{C} = \{C_j\}, j \in (1, m)$, and 
    $C_j$ can be divided into intrinsic parameters $K_j$ and extrinsic parameters $T_j$
    \Ensure Corresponding re-projected 2d coordinates 
    \State Transform 3D points in world frame into camera frame, $X_{ij}^{camera} = T_jX_i^{world}$
    or $X_{ij}^{camera} = R_jX_i^{world} + t_j$
    \State Project the 3D points in camera frame into normalized image plane and get $u_{ij} = (x_{ij}, y_{ij})^T$
    \State Do distortion correction for the 2D points in normalized image plane by (\ref{radial_distort_eqa}) and (\ref{tangential_distort_eqa})
    %\begin{equation}
    %    \begin{aligned}
    %        x_{ij}^c = x_{ij}(1 + k_1r^2 + k_2r^4 + k_3r^6) + 2p_1x_{ij}y_{ij} + p_2(r^2 + 2x_{ij}^2)\\
    %        y_{ij}^c = y_{ij}(1 + k_1r^2 + k_2r^4 + k_3r^6) + p_1(r^2 + 2y_{ij}^2) + 2p_2x_{ij}y_{ij}
    %    \end{aligned}
    %\end{equation}
    \State Project the corrected points into pixel plane by intrinsic parameters
    \begin{equation}
        \left\{
        \begin{aligned}
            \mu_{ij} = f_xx_{ij}^c + c_x\\
            \nu_{ij} = f_yy_{ij}^c + c_y
        \end{aligned}
        \right.
    \end{equation}
    or in matrix representation
    \begin{equation}
        \left[
        \begin{array}{cc}
             \textbf{p} & 1 
        \end{array}
        \right]^T
        = 
        \left[
        \begin{array}{ccc}
             f_x & s & c_x \\
             0 & f_y & c_y \\
             0 & 0 & 1
        \end{array}
        \right]
        \left[
        \begin{array}{c}
             x_{ij}^c \\
             y_{ij}^c \\
             1
        \end{array}
        \right]
    \end{equation}
\end{algorithmic}
\end{algorithm}

% end of Projection Camera Model

\section{Conventional Bundle Adjustment}
\begin{figure}
    \centering
    \includegraphics[scale=0.33]{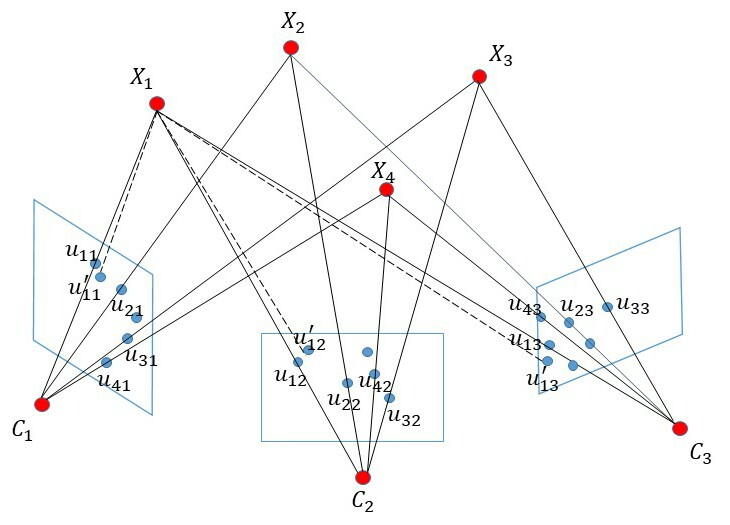}
    \caption{Bundle Adjustment: $u_{ij}$ is the observations, $X_i$ is the 3D points in world frame, $C_j$ is the camera center, $u_{ij}^{'}$ is the re-projected 2D points. The solid line represents the projection procedure, the dotted line represents the re-projection procedure.}
    \label{fig:bundle_adjustment}
\end{figure}

Bundle adjustment is commonly used in structure from motion and in the back end of SLAM. It tries to minimize the sum of errors between 2D observations and the predicted 2D points, where the predicted points are re-projected from 3D structures by camera parameters. It measures the accuracy of the computed 3D structures and camera parameters. As shown in Figure.\ref{fig:bundle_adjustment}, bundle adjustment is actually a nonlinear least square problem, which described by the following equation:
\begin{equation}
\label{std_ba}
    min\ \sum_{i=1}^{n} \sum_{j=1}^m (u_{ij} - \pi(C_j, X_i))^2
\end{equation}
where $u_{ij}$ is the observed point coordinate in pixel level, which represents the $i$th 3D point $X_i$ is observed by the $j$th camera $C_j$. $\pi(C_j, X_i)$ is the nonlinear operation described in section III.\\

%%%%%%%%%%%%%% derivative of Gauss-Newton and Levenberg-Marquardt %%%%%%%%%%%%%%%

To simplify the notation, let $r_{ij} = u_{ij} - \pi(C_j, X_i)$, then we can rewrite (\ref{std_ba}) by
\begin{equation}
\label{sba_short}
    min\ r^Tr
\end{equation}
then, we perform first order Taylor expansion
\begin{equation}
\label{1st_taylor_expansion}
    r(x + \delta x) = r(x) + g^T\delta x + \frac{1}{2}\delta x^T H \delta x
\end{equation}
where $g$ and $H$ is the gradient and Hessian of $r$, respectively. By taking the derivative of equation\ref{1st_taylor_expansion} and setting it to zero, we obtain
\begin{equation}
\label{GN}
    H\delta x = -g
\end{equation}

For least square problems, $H = J^TJ + S, g = J^Tr$, where $S = \sum_{i=1}^n \sum_{j=1}^m r_{ij} \nabla^2 r_{ij} $. If $S$ is small enough, then by omitting $S$, we obtain the Gauss-Newton equation
\begin{equation}
\label{LM}
    J^TJ\delta x = -J^Tr
\end{equation}

Note that, the Gauss-Newton method can always produce a decrease direction if $J^TJ$ is positive definite, because $(\delta x)^Tg = -(\delta x)^TJ^Tr = -(\delta x)^TJ^TJ\delta x < 0$. But when $J^TJ$ becomes singular, then the Gauss-Newton method becomes numerically unstable. To overcome the weakness of Gauss-Newton, we can use the Levenberg-Marquardt method instead.
\begin{equation}
    (J^TJ + \lambda I)\delta x = -J^Tr
\end{equation}

Since the non-linear least square problem has been well studied, we can easily solve the bundle adjustment problem with (\ref{GN}) or (\ref{LM}). In addition, the residual error vector $r$ can be defined in such a way that the square sum represents a robust cost function, rather than just an outlier-sensitive plain least squares cost function\cite{Engels06bundleadjustment}. With a robustifier $\Lambda_x$, we rewrite (\ref{sba_short}) into:
\begin{equation}
    min\ r^T\Lambda_x r
\end{equation}
for (\ref{GN}):
\begin{equation}
\label{normal_equation}
    J^T\Lambda_x J \delta x = -J^T\Lambda_x r
\end{equation}
and for (\ref{LM}):
\begin{equation}
\label{augment_normal_equation}
    (J^T\Lambda_x J + \lambda I)\delta x = -J^T\Lambda_x r
\end{equation}

(\ref{normal_equation}) and (\ref{augment_normal_equation}) are called \textit{Normal Equation} and \textit{Augmented Normal Equation}, respectively.

\subsection{Reduced Camera System}
It seems without any consideration, we can solve (\ref{normal_equation}) and (\ref{augment_normal_equation}) directly by linear algebra. However, for a large scale problem, i.e. we have about 1,000 cameras and 2,000,000 3D points, it becomes impractical to solve them directly for memory limits and efficiency consideration.\\  
For convenience, we use (\ref{normal_equation}) to show how to solve the bundle adjustment problem.
set $\hat{u}_{ij} = \pi(C_j, X_i)$, and we order the parameter $x$ into camera block $c$ and structure block $p$:
\begin{equation}
    x = [c,\ p]
\end{equation}
it's easily to realize that:
\begin{equation}
    J_{ij} = \frac{\partial r_{ij}}{\partial x_k} = \frac{\partial \hat{u}_{ij}}{\partial x_k},\\
    \frac{\partial \hat{u}_{ij}}{\partial c_k} = 0, \forall j \ne k,\\
    \frac{\partial \hat{u}_{ij}}{\partial p_k} = 0, \forall i \ne k
\end{equation}
Consider now, that we have $m = 3$ cameras and $n = 4$ 3D points. Set $A_{ij} = \frac{\partial \hat{u_{ij}} }{\partial c_j}, B_{ij} = \frac{\partial \hat{u_{ij}} }{\partial p_i}$, we can obtain the Jacobi:
\begin{equation}
    J = \frac{\partial \hat{u}}{\partial x}
    =\left[ \begin{array}{ccccccc}
    A_{11} & 0 & 0 & B_{11} & 0 & 0 & 0 \\
    0 & A_{12} & 0 & B_{12} & 0 & 0 & 0 \\
    0 & 0 & A_{13} & B_{13} & 0 & 0 & 0 \\
    A_{21} & 0 & 0 & 0 & B_{21} & 0 & 0 \\
    0 & A_{22} & 0 & 0 & B_{22} & 0 & 0 \\
    0 & 0 & A_{23} & 0 & B_{23} & 0 & 0 \\
    A_{31} & 0 & 0 & 0 & 0 & B_{31} & 0 \\
    0 & A_{32} & 0 & 0 & 0 & B_{32} & 0 \\
    0 & 0 & A_{33} & 0 & 0 & B_{33} & 0 \\
    A_{41} & 0 & 0 & 0 & 0 & 0 & B_{41} \\
    0 & A_{42} & 0 & 0 & 0 & 0 & B_{42} \\
    0 & 0 & A_{43} & 0 & 0 & 0 & B_{43} 
    \end{array} 
    \right ]
\end{equation}
It's very clear that the Jacobi is a very sparse matrix, it's a very important property of which we should make good use. Let's go a step further. 
Set the robustifier becomes block diagonal matrix $diag\{\Lambda_{u_{11}}, \Lambda_{u_{12}}, ..., \Lambda_{u_{nm}},\}$, and
\begin{equation}
    U_j = \sum_{i=1}^4 A_{ij}^T \Lambda_{u_{ij}} A_{ij}, \\
    V_i = \sum_{j=1}^3 B_{ij}^T \Lambda_{u_{ij}} B_{ij}, \\
    W_{ij} = A_{ij}^T \Lambda_{u_{ij}} B_{ij}
\end{equation}
then the left side of (\ref{normal_equation}) becomes:
\begin{equation}
\label{lne}
    J^T \Lambda_u J = \left[ \begin{array}{ccccccc}
         U_1 & 0 & 0 & W_{11} & W_{21} & W_{31} & W_{41} \\
         0 & U_2 & 0 & W_{12} & W_{22} & W_{32} & W_{42} \\
         0 & 0 & U_3 & W_{13} & W_{23} & W_{33} & W_{43} \\
         W_{11}^T & W_{12}^T & W_{13}^T & V_1 & 0 & 0 & 0 \\
         W_{21}^T & W_{22}^T & W_{23}^T & 0 & V_2 & 0 & 0 \\
         W_{31}^T & W_{32}^T & W_{33}^T & 0 & 0 & V_3 & 0 \\
         W_{41}^T & W_{42}^T & W_{43}^T & 0 & 0 & 0 & V_4
    \end{array}
    \right]
\end{equation}

and the right side of (\ref{normal_equation}) becomes:
\begin{equation}
\label{rne}
\begin{aligned}
    J^T \Lambda r = 
    (
        \sum_{i=1}^4(A_{i1}^T \Lambda_{u_{i1}} r_{i1})^T,\  \sum_{i=1}^4(A_{i2}^T \Lambda_{u_{i2}} r_{i2})^T, \\ \sum_{i=1}^4(A_{i3}^T \Lambda_{u_{i3}} r_{i3})^T,\ \sum_{j=1}^3 (B_{ij}^T \Lambda_{u_{1j}} r_{ij})^T, \\ \sum_{j=1}^3 (B_{ij}^T \Lambda_{u_{2j}} r_{2j})^T,\ \sum_{j=1}^3 (B_{ij}^T \Lambda_{u_{3j}} r_{3j})^T,\\ \sum_{j=1}^3 (B_{ij}^T \Lambda_{u_{4j}} r_{4j})^T
    )^T
\end{aligned}
\end{equation}
Set
\begin{equation}
\label{detail_r}
    r_{c_j} = \sum_{i=1}^4(A_{ij}^T \Lambda_{u_{ij}} r_{ij})^T, r_{p_i} = \sum_{j=1}^3(B_{ij}^T \Lambda_{u_{ij}} r_{ij})^T,
r_{ij} = u_{ij} - \hat{u_{ij}}
\end{equation}
and
\begin{equation}
\label{brief_UVW}
\begin{aligned}
    U = diag\{U_1, U_2, U_3\}, \\
    V = diag\{V_1, V_2, V_3, V_4\}, \\
    W = \left[ 
    \begin{array}{cccc}
         W_{11} & W_{21} & W_{31} & W_{41} \\
         W_{12} & W_{22} & W_{32} & W_{42} \\
         W_{13} & W_{23} & W_{33} & W_{43} 
    \end{array}
    \right]
\end{aligned}
\end{equation}

Substitute (\ref{detail_r}) and (\ref{brief_UVW}) to (\ref{lne}) and (\ref{rne}), respectively, we can obtain
\begin{equation}
    J^T \Lambda_u r = [r_{c_1}, r_{c_2}, r_{c_3}, r_{p_1}, r_{p_2}, r_{p_3}, r_{p_4}]^T = [r_c, r_p]^T
\end{equation}
and
\begin{equation}
\label{block_ne}
    J^T \Lambda_u J = 
    \left[
    \begin{array}{cc}
        U & W  \\
        W^T & V 
    \end{array}
    \right]
\end{equation}
Then, we set 
\begin{equation}
\label{block_delta_x}
    \delta x = [\delta c, \delta p]^T
\end{equation}
Substitute (\ref{block_delta_x}) and (\ref{block_ne}) into (\ref{normal_equation}), we obtain
\begin{equation}
\label{final_block_ne}
    \left[
    \begin{array}{cc}
        U & W  \\
        W^T & V
    \end{array}
    \right]
    \left[
    \begin{array}{c}
         \delta c \\
         \delta p 
    \end{array}
    \right] = 
    \left[
    \begin{array}{c}
         r_c \\
         r_p 
    \end{array}
    \right]
\end{equation}
Realize that (\ref{final_block_ne}) becomes an medium to large scale linear equation, if we have more cameras and 3D points. Then to solve it efficiently, we need more tricks. Consider that the number of cameras $m$ far less than the number of 3D points $n$, we can eliminate structure block $\delta p$ and get solution of $\delta c$, then we obtain the result of $\delta p$ by back-substitution. Now left multiply $[I -WV^{-1}]^T$ on both side of (\ref{final_block_ne}), then we can get
\begin{equation}
\label{el_RCS}
    \left[
    \begin{array}{cc}
         U-WV^{-1}W^T & 0 
    \end{array}
    \right]
    \left[
    \begin{array}{c}
         \delta c \\
         \delta p 
    \end{array}
    \right] =\
    \left[
    \begin{array}{cc}
         I & -WV^{-1}  \\
    \end{array}
    \right]
    \left[
    \begin{array}{c}
         r_c \\
         r_p
    \end{array}
    \right]
\end{equation}
then, from (\ref{el_RCS}) we further get
\begin{equation}
\label{RCS}
    (U - WV^{-1}W^T)\delta c = r_c - WV^{-1}r_p
\end{equation}
(\ref{RCS}) is called the \textit{Reduced Camera System}, and $S = U - WV^{-1}W^T$ is called the \textit{Schur Complement} of $V$ in the left-hand side of (\ref{final_block_ne}). It can prove that the schur complement of a symmetric positive definite matrix is symmetric positive definite, thus (\ref{RCS}) can be solved by Cholesky decomposition.\\
With (\ref{el_RCS}), we have 
\begin{equation}
\label{no_name}
    W^T\delta c + V\delta p = r_c
\end{equation}
then, by solving (\ref{no_name}), we can obtain
\begin{equation}
\label{delta_p}
    \delta p = V^{-1}(r_p - W^T\delta c)
\end{equation}

Then back to the little example such that $m = 3$ and $n = 4$, we can obtain
\begin{equation}
\label{Schur complement}
    S = % U_{3\times3}-W_{3\times4}V_{4\times4}^{-1}W_{4\times3}^T = 
    \left[
    \begin{array}{lll}
         U_1 - \sum_{i=1}^4\limits Y_{i1}W_{i1}^T & -\sum_{i=1}^4\limits Y_{i1}W_{i2}^T & -\sum_{i=1}^4\limits Y_{i1}W_{i3}^T \\
         -\sum_{i=1}^4\limits Y_{i2}W_{i1}^T & U_2 - \sum_{i=1}^4\limits Y_{i2}W_{i2}^T & -\sum_{i=1}^4\limits U_{i2}W_{i3}^T \\ 
         -\sum_{i=1}^4\limits Y_{i3}W_{i1}^T & -\sum_{i=1}^4\limits Y_{i3}W_{i2}^T & U_3 - \sum_{i=1}^4\limits Y_{i3}W_{i3}^T
    \end{array}
    \right]
\end{equation}
By setting $Y_{ij} = W_{ij}V_I^{-1}$, we can rewrite the right side of (\ref{RCS})
\begin{equation}
\label{r_rcs}
    r_c - WV^{-1}r_p = 
    \left[
    \begin{array}{c}
         r_{a_1}-\sum_{i=1}^4(Y_{i1}r_{b_i})^T \\ 
         r_{a_2}-\sum_{i=1}^4(Y_{i2}r_{b_i})^T \\ 
         r_{a_3}-\sum_{i=1}^4(Y_{i3}r_{b_i})^T
    \end{array}
    \right]
\end{equation}
By inserting (\ref{Schur complement}) and (\ref{r_rcs}) into (\ref{RCS}), we could solve the reduced camera system easily, and get the solution of (\ref{delta_p})
\begin{equation}
    \delta p = V^{-1}(r_p-W^T\delta c) = 
    \left[
    \begin{array}{c}
         V_1^{-1}(r_{b_1} - \sum_{j=1}^3W_{1j}^T\delta c_j) \\
         V_2^{-1}(r_{b_2} - \sum_{j=1}^3W_{2j}^T\delta c_j) \\
         V_3^{-1}(r_{b_3} - \sum_{j=1}^3W_{3j}^T\delta c_j) \\
         V_4^{-1}(r_{b_4} - \sum_{j=1}^3W_{4j}^T\delta c_j)
    \end{array}
    \right]
\end{equation}

Note that, the derivation of the reduced camera system is based on the normal equation(\ref{normal_equation}), which suggests that we are using Gauss-Newton method. If we want to use Levenber-Marquardt method, we should make some modification of the equations above based on the augmented normal equation. As we just need to augment the diagonal elements of $U$ and $V$, then we should only replace $U$ and $V$ by $U^*$ and $V^*$, respectively, where $*$ represents the augmentation of the diagonal elements of $U$ and $V$. Besides, if a 3D point $X_k$ is not observed by camera $l$, then $A_{kl} = \frac{\partial \hat{u}_{kl}}{\partial p_k} = 0, B_{kl} = \frac{\partial \hat{u}_{kl}}{\partial c_l} = 0$.\\

With the derivation above, we can extend it into any scale problems such that $m=M, n=N$ and lead to the algorithm below for solving the (augmented) normal equation.
\begin{algorithm}
\caption{Solving the (Augmented) Normal Equation}
\label{ane_alg}
\begin{algorithmic}[1]

\Require
2-dimensional pixel coordinates of measured data $\boldsymbol{u_{ij}}$,
$m$ cameras parameters $\boldsymbol{c_j}$,
$n$ observed 3D points coordinates $\boldsymbol{X_i}$,
Nonlinear function $\pi(\boldsymbol{c_j},\ \boldsymbol{X_i})$
\Ensure
The increment $\delta p_i$,
The increment $\delta c_j$

\State Compute the derivation matrix: 
$$A_{ij} = \frac{\partial \hat{u_{ij}}}{\partial c_j},\ B_{ij} = \frac{\partial \hat{u_{ij}}}{\partial X_i}$$

$$r_{ij} = u_{ij} - \pi(c_j,\ X_i) = u_{ij} - \hat{u_{ij}}$$

\State Compute the temporal variables:
$$U_j = \sum_{i=1}^n A_{ij}^T \Lambda_{u_{ij}} A_{ij},\ V_i = \sum_{j=1}^m B_{ij}^T \Lambda_{u_{ij}} B_{ij}$$ 
$$ W_{ij} = A_{ij}^T \Lambda_{u_{ij}}B_{ij} \qquad \qquad \qquad \qquad \qquad \quad $$
$$r_{c_j} = \sum_{i=1}^n A_{ij}^T \Lambda_{u_{ij}} r_{ij},\  r_{p_i} = \sum_{j=1}^m B_{ij}^T \Lambda_{u_{ij}} r_{ij}$$

\State 
(Optional) if using Levenberg-Marquardt method, augment $U$ and $V$
$$U_j^* = (\sum_{i=1}^n A_{ij}^T \Lambda_{u_{ij}} A_{ij} ) + \lambda,\  V_i^* = (\sum_{j=1}^m B_{ij}^T \Lambda_{u_{ij}} B_{ij} ) + \lambda $$

\State
Compute the temporal variables
$$T_{ij} = W_{ij}V_i^{-1} \qquad \qquad \quad$$
$$S_{jk} = \mu_{jk}U_j - \sum_{i=1}^n Y_{ij}W_{ik}^T $$
$$r_j = r_{c_j} - \sum_{i=1}^n (Y_{i1}r_{p_i})^T \quad$$
where $\mu_{ik}$ is the Kronecker's delta

\State Solve equation $S\cdot[\delta c_1,\dots,\delta c_m]^T = [r_1,\dots, \delta c_m]^T$

\State Compute $\delta p_i = V_i^{-1}(r_{p_i} - \sum_{j=1}^mW_{ij}^T\delta c_j$
\end{algorithmic}
\end{algorithm}

\subsection{Bundle Adjustment with Conjugate Gradient}
The algorithm.\ref{ane_alg} that solve the (augmented) normal equation is used by \cite{DBLP:journals/toms/LourakisA09}, and \cite{DBLP:journals/toms/LourakisA09} uses the \textit{Compressed Row Storage(CRS)} format for storing the sparse Jacobi. However, \cite{DBLP:journals/toms/LourakisA09} needs to store Jacobi and $S$ explicitly, and the dense decomposition of $S$, make it only suitable for small to medium scale problems. Besides, due to the noise of observed data and the loss of accuracy during computation, the condition number of $J^TJ$(or $J^T \Lambda J$) becomes large, then the Gauss-Newton and Levenberg-Marquardt methods converge slowly. Then, some works try to seek approaches to overcome it, and preconditioned conjugate gradient is considered commonly \cite{DBLP:journals/cacm/AgarwalFSSCSS11, DBLP:conf/eccv/AgarwalSSS10, DBLP:conf/eccv/ByrodA10, DBLP:conf/bmvc/ByrodA09, DBLP:conf/iccv/JianBD11}. 

\subsubsection{Standard Conjugate Gradient Algorithm}
The conjugate gradient algorithm is an iterative method for solving a symmetric positive definite system of linear equations
\begin{equation}
\label{linear_equation}
    Ax = b
\end{equation}
It's equivalent to solve the quadratic equation
\begin{equation}
\label{quadratic_equation}
    min\ q(x) = \frac{1}{2}x^TAx - b^Tx
\end{equation}
The standard conjugate gradient algorithm for solving (\ref{linear_equation}) is given in algorithm\ref{std_cg_alg}.
\begin{algorithm}
\caption{Standard Conjugate Gradient Algorithm}
\label{std_cg_alg}
\begin{algorithmic}[1]
\Require
$x_0$, $A$, $b$
\Ensure
$x^*$
\State 
$s^0 = b - Ax^0, p^0 = s^0, k = 0$
\While{$\lVert s^k\rVert > threshold $}
    \State $\alpha^k = \frac{(s^k)^Ts^k}{(p^k)^TAp^k}$
    \State $x^{k+1} = x^k + \alpha^kp^k$
    \State $s^{k=1} = s^k - \alpha^kAp^k$
    \State $\beta^k = \frac{(s^{k+1})^Ts^{k+1}}{(s^k)^Ts^k}$
    \State $p^{k+1} = s^{k+1} + \beta^kp^k$
    \State k = k + 1
\EndWhile
\end{algorithmic}
\end{algorithm}

\subsubsection{Inexact Newton Methods}
Since the Newton or Gauss-Newton or Levenberg-Marquardt step is an approximation to the true optimum, there is no need to solve the normal equations very exactly and it's likely to be a good idea to abort the linear conjugate gradient method early, going for an approximate solution.\\
A typical termination is of the form
\begin{equation}
    \lVert H\delta x + g \rVert \leq \eta_j \lVert g \rVert
\end{equation}
$\eta_j \in (0,1)$ is called \textit{forcing sequence}, $\eta_j  = 0.1$ is suggested in \cite{DBLP:conf/eccv/ByrodA10}.

\subsubsection{Preconditioned Conjugate Gradient}
Given a linear system (\ref{linear_equation}) and a preconditioner $M$, the preconditioned system is given by 
\begin{equation}
\label{preconditioned_system}
    M^{-1}Ax = M^{-1}b
\end{equation}
For each iteration of PCG to be efficient, $M$ should be cheaply invertible and for the number of iterations of PCG to be small, the condition number $\kappa(M^{-1}A)$ should be as small as possible.
The ideal preconditioner would be one for which $\kappa(M^{-1}A = 1$. The simplest of all preconditioners is the diagonal or Jacobi preconditioner, $M = diag(A)$. For $H$ in (\ref{GN}), it has the property that its diagonal blocks $A$ and $B$ are themselves block diagonal matrices, this property makes the block \textit{Jacobi preconditioner} 
\begin{equation}
    M_J = 
    \left[
    \begin{array}{cc}
         A & 0 \\
         0 & B
    \end{array}
    \right]
\end{equation}
the optimal diagonal preconditioner for $H$.

Some approaches apply PCG to the reduced camera matrix $S$ instead of $H$, because $S$ is a much smaller matrix than $H$, and $\kappa(S) \leq \kappa(H)$. There are two obvious choices for block diagonal preconditioners for $S$, matrix $A$ and block diagonal $\mathcal{D(S)}$, the block Jacobi preconditioner for $S$.
Consider the generalized \textit{Symmetric Successive Over Relaxation(SSOR)} preconditioner for $H$,
\begin{equation}
    M_W(P) = 
    \left[ \begin{array}{cc}
         P & \omega W \\
         0 & B
    \end{array} \right]
    \left[ \begin{array}{cc}
         P^{-1} & 0 \\
         0 & B^{-1} 
    \end{array} \right]
    \left[
    \begin{array}{cc}
         P & 0 \\
         \omega W^T & B 
    \end{array}
    \right]
\end{equation}
where $P$ is some easily invertible matrix and $0 \leq \omega < 2$ is a scalar parameter.

Note that, for $\omega = 0$, $M_0(B) = M_J$ is the block Jacobi preconditioner. For $\omega = 1$, it can be shown that using $M_1(P)$ as a preconditioner for $H$ is exactly equivalent to using the matrix $P$ as a preconditioner for the reduced camera matrix $S$. This means that for $P = I$ using $M_1(I)$ as a precondioner for $H$ is the same as running pure conjugate gradient on $S$ and we can run PCG on $S$ with preconditioner $A$ and $\mathcal{D}(S)$ by using $M_1(A)$ and $M_1(\mathcal{D}(S))$ as preconditioner for $H$. Thus, the schur complement which started out its life as a way of specifying the order in which the variables should be eliminated from $H$ when solving \ref{GN} exactly, returns to the scene as a generalized SSOR preconditioner when solving the same linear system iteratively.

In the work of \cite{DBLP:conf/eccv/AgarwalSSS10}, they use six kinds of bundle adjustment alogrithms:
\begin{itemize}
    \item \textit{Explicit-Direct}, which explicity construct the schur complement matrix $S$ and solve (\ref{el_RCS}) using dense factorization. This leads to the work of \cite{DBLP:journals/toms/LourakisA09}.
    \item \textit{Explicit-Sparse}, which explicity construct the schur complement matrix $S$ and solve (\ref{el_RCS}) using sparse direct factorization.
    \item \textit{Explicit-Jacobi}, which explicity construct the schur complement matrix $S$ and solve (\ref{el_RCS}) using PCG, which uses the block Jacobi preconditioner $\mathcal{D}(S)$.
    \item \textit{Normal-Jacobi}, which uses PCS on $H$ with the block Jacobi pre-conditioner $\mathcal{D}(S)$.
    \item \textit{Implicit-Jacobi}, which runs PCG on $S$ using block Jacobi preconditioner $\mathcal{D}(S)$.
    \item \textit{Implicit-SSOR}, which runs PCG on $S$ using the block Jacobi preconditioner $A$.
\end{itemize}
As \cite{DBLP:conf/eccv/AgarwalSSS10} summarized, for small to medium problems, the use of a dense Cholesky-based LM algorithm is recommended. For larger problems, the situation is more complicated and there is no one clear answer. Both Implicit-SSOR and Explicit-Jacobi offer competitive solvers, with Implicit-SSOR being performed for problems with lower sparsity and Explicit-Jacobi for problems with high sparsity. And once the cuase of numerical instability in Implicit-Jacobi can be understood and rectified, it will offer a memory efficient solver that bridges the gap between these two solvers and works on large bundle adjustment problems, independent of their sparsity.

\subsubsection{Block QR Preconditioning}
As mentioned in the last section, we talk about the preconditioned conjugate gradient briefly. However, there are more tricks of the preconditioning\cite{DBLP:conf/eccv/ByrodA10}.

In the case of the conjugate gradient method means pre-multiplying from left and right with a matrix $E$ to form
\begin{equation}
    E^TAE\hat{x} = E^Tb
\end{equation}
where $E$ is a non-singular matrix.

The idea is to select $E$ so that $\hat{A} = E^TAE$ has a smaller condition number than $A$. Finally, $x$ can be computed from $\hat{x}$ with $x = E\hat{x}$. Often $E$ is chosen so that $EE^T$ approximates $A^{-1}$ in some sense.

Explicitly forming $\hat{A}$ is expensive and usually avoided by inserting $M= EE^T$ in the right places in the conjugate gradient method obtaining the preconditioned conjugate gradient method. Two useful preconditioners can be obtained by writting $A = L + L^T - D$, where $D$ and $L$ are the diagonal and lower triangular parts of $A$. Setting $M=D^{-1}$ is known as \textit{Jacobi Preconditioning} and $M = L^{-T}DL^{-1}$ is \textit{Gauss-Seidel Preconditioning}.

The Jacobi and Gauss-Seidel preconditioners do not make use of the special structure of the bundle adjustment Jacobian. Assume we have the QR decomposition of $J$, $J = QR$ and set $E = R^{-1}$. This yields the preconditioned normal equation
\begin{equation}
    R^{-T}J^TJR^{-1}\delta\hat{x} = -R^{-T}J^Tr
\end{equation}
then we can obtain
\begin{equation}
    R^{-T}(QR)^T(QR)R^{-1}\delta \hat{x} = -R^{-T}J^Tr
\end{equation}
and we can get the solution of $\delta \hat{x}$
\begin{equation}
    \delta \hat{x} = -R^{-T}J^Tr
\end{equation}

However, computing $J = QR$ is an expensive operation. Then, consider the Jacobi partitioning $J = [J_c, J_p$, we can do a block-wise factorization:
\begin{equation}
    J_c = Q_cR_c, J_p = Q_pR_p
\end{equation}
Due to the special block structure of $J_c$ and $J_p$, we have
\begin{equation}
\begin{aligned}
    R_c = R(J_c) = diag\{R(A_1), R(A_2), \dots, R(A_n)\},\\
    R_p = R(J_p) = diag\{R(B_1), R(B_2), \dots, R(B_n)\}
\end{aligned}
\end{equation}
where 
$
A_k = 
\left[
\begin{array}{c}
     A_{k1} \\
     A_{k2} \\
     \vdots \\
     A_{kn}
\end{array}
\right]
$
, and 
$
\left[
\begin{array}{c}
     B_{1k} \\
     B_{2k} \\
     \vdots \\
     B_{mk}
\end{array}
\right]
$.

In other words, we perform QR factorization independently on the block columns of $J_c$ and $J_p$, making this operation very efficient and easy to parallelize. The preconditioner thus becomes
\begin{equation}
    E = 
    \left[
    \begin{array}{cc}
         R(J_c)^{-1} & 0  \\
         0 & R(J_p)^{-1}
    \end{array}
    \right]
\end{equation}

QR preconditioner is equivalent to block-Jacobi preconditioning, but:
\begin{itemize}
    \item we do not need to form $J^TJ$
    \item QR factorization of a matrix $A$ is generally considered numerically superior to forming $A^TA$ followed by Cholesky factorization.
\end{itemize}

Consider the preconditioned normal equation:\\
$R^{_T}J^TJR^{-1}\delta x = -R^{-T}J^Tr$
Set $\hat{J} = JR^{-1}$, then it becomes $\hat{J}^T\hat{J}\delta x = -\hat{J}^Tr$, and
\begin{equation}
    \hat{J}^T\hat{J} = 
    % \left[ \begin{array}{cc}
    %         R(J_c) & 0 \\
    %         0 & R(J_p)
    %    \end{array} \right]^{-T}
    % \left[ \begin{array}{cc}
    %         J_c^TJ_c & J_cJ_p  \\
    %          J_p^TJ_c & J_p^TJ_p
    %     \end{array} \right]
    % \left[ \begin{array}{cc}
    %          R(J_c) & 0 \\
    %          0 & R(J_p)
    %     \end{array} \right]^{-1}\\
    % =
    \left[ \begin{array}{cc}
             Q_c^TQ_c & Q_c^TQ_p \\
             Q_p^TQ_c & Q_p^TQ_p 
        \end{array} \right]
\end{equation}
where $Q_c^T$ and $Q_pQ_p$ are both identity matrices and $Q_p^TQ_c = (Q_c^TQ_p)^T$.
Set $s^k = \left[ \begin{array}{c}
     s_c^k \\
     s_p^k 
\end{array} \right]$, applying Reid's\cite{DBLP:conf/eccv/ByrodA10} results yielding the following: By initializing so that $\delta x_c = 0$ and $\delta x_p = -J_p^Tr$, we will have $s_c^{2m} = s_p^{2m+1} = 0$. Then we can extends the standard conjugate algorithm \ref{std_cg_alg} into a block QR factorization conjugate gradient algorithm in the following:
\begin{algorithm}
\caption{Block QR factorization Conjugate Gradient}
\label{block_QR_CG}
\begin{algorithmic}[1]
    \State $\eta = 0.1$
    \State $\delta x_c^0 = 0, \delta x_p^0 = -J_p^Tr, \hat{r}^0 = -r - J(\delta p)^0,
    p^0 = s^0 = J^T\hat{r}^0, \gamma^0 = (s^0)^TS^0, q^0 = Jp^0, k = 0$
    \While{$\lVert S^k \rVert > \eta\lVert S^0 \rVert$}
        \State $\alpha^k = \frac{\gamma^k}{(q^k)^Tq^k}$
        \State $\delta p^{k+1} = \delta p^k + \alpha^k p^k$
        \State $s_c^{k+1} = -alpha^k J_c^Tq^k, s_p^{k+1} = 0$, if $k$ odd 
        \State $s_x^{k+1} = -\alpha^kJ_p^Tq^k, s_c^{k+1} = 0$, if $k$ even
        \State $\gamma^{k+1} = (s^{k+1})^Ts^{k+1}$
        \State $\beta^k = \frac{\gamma^{k+1}}{\gamma^k}$
        \State $p^{k+1} = s^{k+1} + \beta^kp^k$
        \State $q^{k+1} = \beta^kq^k + J_cs_c^{k+1}$, if $k$ odd,
        \State $q^{k+1} = \beta^kq^k + J_ps_p^{k+1}$, if $k$ even
    \EndWhile
\end{algorithmic}
\end{algorithm}
One further interesting aspect of matrices is that one can show that for these matrices like (\ref{block_ne}), block-Jacobi preconditioning is always superior to Gauss-Seidel and SSOR preconditioners.

%\subsection{Bundle adjustment in Parallel}
%Though bundle adjustment is well studied until now, demands for very large scale reconstruction are increased. The most exciting works of city-scale reconstructions are \textbf{TODO: HKUST}. Due to the limitation of computer memory, people is not satisfied with just complement the bundle adjustment in just one single computer. 

% end of Bundle Adjustment

\section{Distributed Bundle Adjustment}
% Though bundle adjustment is well studied until now, demands for very large scale reconstruction are increased. The most exciting works of city-scale reconstructions are \textbf{TODO: HKUST}. Due to the limitation of computer memory, people is not satisfied with just complement the bundle adjustment in just one single computer.

The conventional bundle adjustment that based on the Levenber-Marquardt algorithm is limited to single machine, thus when the problem gets larger and larger, the conventional approaches suffer from memory limitation and have to afford heavy computation burden. \cite{DBLP:conf/cvpr/ErikssonBCI16} is the first that proposed to perform bundle adjustment in distributed manner. 

Starting from a general convex optimization problem:
\begin{equation}
\label{general_convex_optimization}
    min_{x\in H}\ \sum_{i=1}^N f_i(x)
\end{equation}
where $f_i$, $i=1,\dots,N$ are convex and lower semi-continuous functions a $H$ a Hilbert space. By adopting \textit{proximal splitting} methods\cite{DBLP:books/sp/11/CombettesP11, DBLP:conf/scalespace/Setzer09}, we can define the \textit{proximity operator} $prox_f: H \to H$ of a proper, convex and lower semi-continuous function $f: H \to \mathcal{R}$ as 
\begin{equation}
\label{proximal_splitting}
    prox_{f/\rho}(y) = arg_{x\in H}\ min\ (f(x)+ \frac{\rho}{2} \lVert x-y \rVert^2)
\end{equation}
where $\rho > 0$ and $H$ a Hilbert space. By taking the result of \textit{Douglas-Rachford} splitting schemes\cite{DBLP:books/sp/11/CombettesP11}, we can get the fix-point iterations of (\ref{general_convex_optimization}) when $N=2$
\begin{equation}
        z^{t+1} = prox_{f_1/\rho}(x^t),\ \qquad \qquad \qquad \qquad 
\end{equation}
\begin{equation}
        x^{t+1} = x^t - z^{t+1} + prox_{f_2/\rho}(2z^{t+1} - x^t)
\end{equation}

Let $H_1$ and $H_2$ be some partition of the Hilbert space $H$, i.e, $H = H_1 \times H_2$. Then the partial proximity operator $prox_f^{\dagger}: H_2 \to H$, of $f: H \to \mathcal{R}$, is defined as
\begin{equation}
    prox_{f/\rho}^{\dagger}(y) = arg\ min\ (f(x_1, x_2) + \frac{\rho}{2} \lVert x_2 - y \rVert^2
\end{equation}
then the fix-point iterations should be modified accordingly, thus arriving at the \textit{partial Douglas-Rachford} splitting
\begin{equation}
\label{partial_dr_splitting1}
    z^{t+1} = prox_{f_1/\rho}^{\dagger} (x_2^t),\ \qquad \qquad \qquad \qquad 
\end{equation}
\begin{equation}
\label{partial_dr_splitting2}
    x^{t+1} = x^t - z^{t+1} + prox_{f_2/\rho}^{\dagger}(2z_2^{t+1} - x_2^t)
\end{equation}

With the equation of (\ref{partial_dr_splitting1}) and (\ref{partial_dr_splitting2}), we can derive the distributed bundle adjustment algorithm.

We split the $m$ images into $l$ disjoint partitions, and $c_k \subseteq \{1,\dots,m\}, k=1,\dots,l$ with $\cup_k c_k =\{1,\dots,m\}$ and $c_i \cap c_j = \emptyset, \forall i \ne j$, and additional latent variables $X_i^k \in \mathcal{R}^3, i=1,\dots,n, k=1,\dots,l$. Then rewrite (\ref{std_ba}) as
\begin{equation}
    \begin{aligned}
            min\ \sum_{k=1}^l \sum_{i=1}^n \sum_{j \in c_k} w_{ij}\lVert u_{ij} - \pi(P_j, \Bar{X}_i^k) \rVert^2 \\ + \sum_{k_1=1}^{l-1} \sum_{k_2=k_1+1}^l \sum_{i=1}^n l_{\textbf{0}} (\Bar{w}_i^{k_1}\Bar{w}_i^{k_2}(\Bar{X}_i^{k_1} - \Bar{X}_i^{k_2}))
    \end{aligned}
\end{equation}
where the visibility matrix is defined by
\begin{equation}
    \Bar{w}_j^k = 
    \left\{
    \begin{aligned}
        1, \exists i \in c_k, s.t. w_{ij} = 1 \\
        0, otherwise
    \end{aligned}
    \right.
\end{equation}
and the the indicator function defined as
\begin{equation}
    l_{\mathcal{S}}(x) = 
    \left\{
    \begin{aligned}
        \infty, x \notin \mathcal{S},\\
        0, x \in \mathcal{S}
    \end{aligned}
    \right.
\end{equation}
 
Letting 
\begin{equation}
    f_1(P, \Bar{X}) = \sum_{k=1}^l \sum_{i=1}^n \sum_{j \in c_k} w_{ij}\lVert u_{ij} - \pi(P_j, \Bar{X}_i^k) \rVert^2,
\end{equation}
and
\begin{equation}
    f_2(P, \Bar{X}) = \sum_{k_1=1}^{l-1} \sum_{k_2=k_1+1}^l \sum_{i=1}^n l_{\textbf{0}} (\Bar{w}_i^{k_1}\Bar{w}_i^{k_2}(\Bar{X}_i^{k_1} - \Bar{X}_i^{k_2})).
\end{equation}
 
 By applying (\ref{partial_dr_splitting1}) and (\ref{partial_dr_splitting2}) of the partial Douglas-Rachford iteration, we can obtain
 \begin{equation}
 \label{dis_partial1}
     \begin{aligned}
         prox_{f_1/\rho}^{\dagger}(Z) = \qquad \qquad \qquad \qquad \qquad \qquad \qquad \qquad \qquad \quad \\ 
         arg\ min\ \sum_{k=1}^l \sum_{i=1}^n \sum_{j \in c_k} w_{ij}\lVert u_{ij} - \pi(P_j, \Bar{X}_i^k) \rVert^2 + \frac{\rho}{2}\lVert \Bar{X} - Z \rVert_F^2
     \end{aligned}
 \end{equation}
 Since the partitions $c_k$ are disjoint, then we can rewrite (\ref{dis_partial1})
 \begin{equation}
 \label{dis_ba1}
    \begin{aligned}
        \sum_{k=1}^l min\ \sum_{i=1}^n\sum_{j\in c_k} w_{ij}\lVert u_{ij} - \pi(P_i, \Bar{X}_i^k) \rVert^2 \\
         + \frac{\rho}{2}\Bar{w}_i^k\lVert \Bar{X}_i^k - Z_i^k \rVert^2
    \end{aligned}
 \end{equation}
 Note that the equation (\ref{dis_ba1}) has one important property: the inner summand of (\ref{dis_ba1}) is completely independent over $k$ and that each of these subproblem is a total sum-of-squares problem to which a standard bundle adjustment solver is directly applicable. Thus, $prox_{f_1/\rho}^{\dagger}$ can be evaluated by solving $l$ smaller independent SfM problem in parallel.
 
 Moreover, we can obtain 
 \begin{equation}
 \label{dis_partial2}
     \begin{aligned}
         prox_{f_2/\rho}(Z) \qquad \qquad \qquad \qquad \qquad \qquad \qquad \qquad \quad \\
         = arg\ min\ \sum_{k_1=1}^{l-1} \sum_{k_2=k_1+1}^l \sum_{i=1}^n l_{\textbf{0}} (\Bar{w}_i^{k_1}\Bar{w}_i^{k_2}(\Bar{X}_i^{k_1} - \Bar{X}_i^{k_2})) \\+ \frac{\rho}{2}\lVert \Bar{X} - Z \rVert_F^2\\
         = \sum_{i=1}^n\Big(min\  \sum_{k_1=1}^{l-1} \sum_{k_2=k_1+1}^l l_{\textbf{0}} \big(\Bar{w}_i^{k_1}\Bar{w}_i^{k_2}(\Bar{X}_i^{k_1} - \Bar{X}_i^{k_2}) \big) \quad\\ 
         + \frac{\rho}{2} \sum_{k=1}^l(\Bar{X}_i^k - Z_i^k)^2 \Big)
     \end{aligned}
 \end{equation}
 and we can find (\ref{dis_partial2}) is independent of $i$, then (\ref{dis_partial2}) can solved in distributed manner. By taking the result of \cite{DBLP:conf/cvpr/ErikssonBCI16}, we can rewrite (\ref{dis_partial2}) as following equation
 \begin{equation}
 \label{dis_ba2}
     [prox_{f_2/\rho}(Z)]_i^k = 
     \left\{
     \begin{aligned}
        \frac{\sum_{k=1}^l \Bar{w}_i^kz_i^k}{\sum_{k=1}^l \Bar{w}_j^k}, \Bar{w}_j^k = 1,\\
        z_i^k, otherwise
     \end{aligned}
     \right.
 \end{equation}
 The above equation of distributed bundle adjustment can be summarized in algorithm \ref{alg: distributed_ba}.
 \begin{algorithm}
 \caption{Distributed Bundle Adjustment}
 \label{alg: distributed_ba}
 \begin{algorithmic}[1]
     \Require image observations $u = \{u_{ij} |\ i \in [1,n], j \in [1, m]\},$ 
     proximal weighting $\{ \rho^{(t)}\}_{t=0}^{\infty}, \rho^{(t)} \in \mathcal{R}^{+}$, 
     camera partitions $c = \{c_k |\ k \in [1, l]\}$, initial camera parameters $P^{(0)}$, initial 3D points $X^{(0)}$ and initial latent variables $\Bar{X}^{k(0)} \gets X^{(0)} $ 
     \State t = 0
     \While{!convergence}
        \State Solve $\{P^{(T+1)}, \Bar{X}^{(t+1)}\} \gets prox^{\dagger}_{f_1/\rho^{(t)}}(Z^{(t)})$,
        and perform $\{P_{j \in c_k}^{(T+1)}, \Bar{X}^{k(t+1)}\} \gets prox^{\dagger}_{f_1/\rho^{(t)}}(Z^{k(t)})$
        for all $k \in [1, l]$ as in (\ref{dis_ba1})
        \State $\{Z^{(t+1)}\} \gets Z^{(t)} - \Bar{X}^{k(t+1)} + prox_{f_2/\rho}(t) (2\Bar{X}^{(t+1)}) - Z^{(t)}$, $prox_{f_2/\rho}$ is given by (\ref{dis_ba2}).
        \State $X_i^{(t+1)} \gets \Bar{X}_i^{k(t+1)}$, for $\forall k \in [1, l]$ such that $\Bar{w}_i^k = 1, i \in [1, n]$
        \State t = t+1
     \EndWhile
 \end{algorithmic}
 \end{algorithm}
 
\section{Conclusion}
In this paper, we make a detailed derivation and kinds of solutions of the bundle adjustment problem, both in a conventional manner and distributed manner. It is clear that preconditioned conjugate gradient methods can replace the dense Cholesky decomposition based method\cite{DBLP:journals/toms/LourakisA09}. While distributed bundle adjustment approaches are prone to solve very large-scale reconstruction problems, they prove surpassing the conventional approaches in small to large-scale problems in both accuracy and efficiency.


\section*{Acknowledgment}
% optional entry into table of contents (if used)
%\addcontentsline{toc}{section}{Acknowledgment}

This work is supported by The National Key Technology Research and Development Program of China under Grants 2017YFB1002705 and 2017YFB1002601, and by National Natural Science Foundation of China(NSFC) under Grants 61472010, 61632003, 61631001 and 61661146002, and by Science and Technology on Complex Electronic System Simulation Laboratory under Grants DXZT-JC-ZZ-2015-019.

\newpage



% http://www.ctan.org/tex-archive/macros/latex/contrib/supported/IEEEtran/testflow/bibtex
\bibliographystyle{ieee.bst}
\bibliography{ref}

\end{document}